\documentclass{vgtc}                          % final (conference style)
\graphicspath{{figures/}{pictures/}{images/}{./}} % where to search for the images

\usepackage{times}                     % we use Times as the main font
\usepackage{tabu}                      % only used for the table example
\usepackage{booktabs}                  % only used for the table example
\usepackage{lipsum}                    % used to generate placeholder text
\usepackage{mwe}                       % used to generate placeholder figures
\usepackage{multirow}
\usepackage{mathptmx}                  % use matching math font
\usepackage{amsmath}

\onlineid{1093}

\vgtccategory{Poster}

\vgtcinsertpkg

\title{GlassesGB: Controllable 2D GAN-Based Eyewear Personalization \\
for 3D Gaussian Blendshapes Head Avatars}

\author{Rui-Yang Ju\thanks{e-mail: jryjry1094791442@gmail.com}\\ %
    \scriptsize Kyoto University
\and Jen-Shiun Chiang\thanks{e-mail: jsken.chiang@gmail.com}\\ %
    \scriptsize Tamkang University}

\teaser{
\setlength{\abovecaptionskip}{-1.5pt}
\setlength{\belowcaptionskip}{-2pt}
\centering
\includegraphics[width=0.98\linewidth]{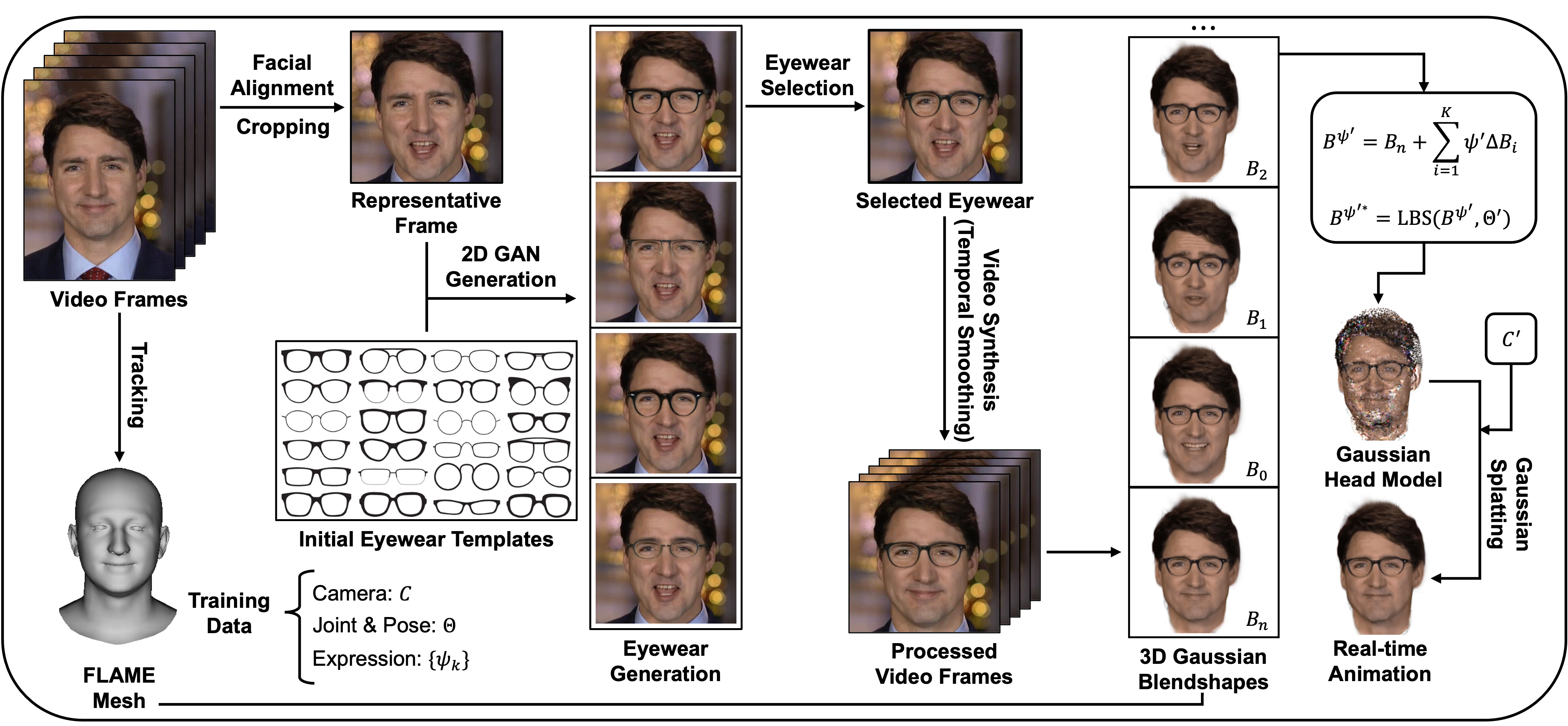}
\caption{We introduce a novel framework, named \textbf{GlassesGB}, that generates personalized eyewear via a 2D GAN-based method and jointly renders it as true 3D geometry on a 3D head avatar using 3D Gaussian Blendshapes, rather than as texture overlays.
}
\label{fig:teaser}
}

\abstract{
Virtual try-on systems allow users to interactively try different products within VR scenarios.
However, most existing VTON methods operate only on predefined eyewear templates and lack support for fine-grained, user-driven customization.
While GlassesGAN enables personalized 2D eyewear design, its capability remains limited to 2D image generation. 
Motivated by the success of 3D Gaussian Blendshapes in head reconstruction, we integrate these two techniques and propose GlassesGB, a framework that supports customizable eyewear generation for 3D head avatars. 
GlassesGB effectively bridges 2D generative customization with 3D head avatar rendering, addressing the challenge in achieving personalized eyewear design for VR applications.
The implementation code is available at
\url{https://ruiyangju.github.io/GlassesGB}.
} % end of abstract


\begin{document}


\firstsection{Introduction}
\maketitle
Compared with traditional physical stores, online shopping offers a significantly more convenient and accessible purchasing experience. 
Eyewear, a daily accessory used by billions of people worldwide, represents a large and rapidly expanding commercial market. 
In this domain, the ability to virtually try on glasses before purchase is crucial for identifying styles that match individual preferences.

With the development of Virtual Reality (VR) technologies, virtual try-on (VTON) has emerged as a solution for interactive product exploration.
By allowing users to preview products within virtual fitting rooms, VTON provides a more informative and engaging shopping experience.
Although existing VTON systems based on 3D modeling can generate visually satisfactory results, they generally support only a predefined set of eyewear types and lack the flexibility required for user-controllable, fine-grained customization.
Therefore, achieving flexible VTON and enabling fully customized eyewear design remains a significant technical challenge.

Generative Adversarial Networks (GANs) have enabled powerful image-editing techniques capable of generating eyewear on facial images. 
Despite the impressive visual quality achieved by existing methods, most still struggle to provide fine-grained parameter control for customizing eyewear appearance. 
GlassesGAN~\cite{plesh2023glassesgan} introduces a GAN-based editing framework that allows users to design personalized eyewear styles and apply them to target faces, offering multiple adjustable parameters for customization. 
However, this method operates only within the 2D image domain, limiting its ability to render side views or novel viewpoints, which is important for users who wish to preview eyewear from different angles.

3D Gaussian Splatting (3DGS)~\cite{kerbl20233d} has demonstrated remarkable efficiency in rendering while preserving high-fidelity geometric details. 
Based on this representation, 3D Gaussian Blendshapes (3DGB)~\cite{ma20243d} integrate 3DGS with expression blendshapes, achieving state-of-the-art (SOTA) performance in 3D head reconstruction. 
However, 3DGB is limited to synthesizing photorealistic avatars that correspond to the input videos and does not support personalized eyewear editing or controllable eyewear synthesis in 3D.

To address this challenge, we propose GlassesGB, a novel framework that integrates the controllable editing capabilities of GlassesGAN with the expressive 3D reconstruction power of 3DGB. 
GlassesGB enables users to design personalized eyewear and render this customized eyewear model together with the target 3D avatar.
To the best of our knowledge, GlassesGB is the first framework that jointly renders controllable GAN-based eyewear as true 3D geometry on 3D Gaussian Blendshapes head avatars.

\begin{figure}[t]
\centering
\includegraphics[width=0.99\columnwidth]{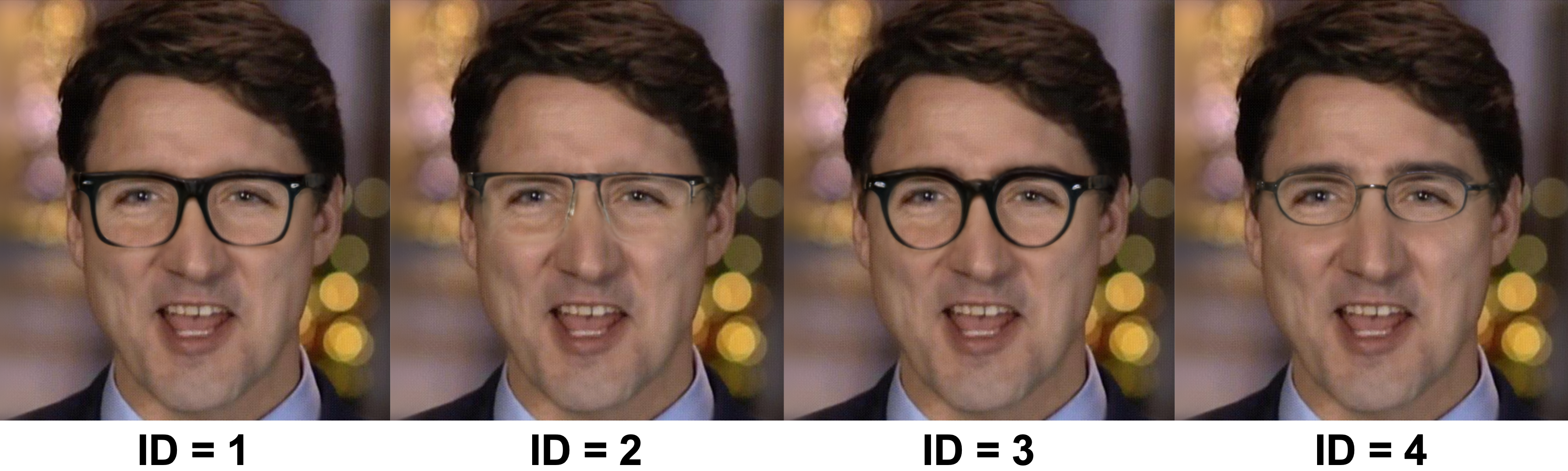}
\resizebox{0.98\linewidth}{!}{
\begin{tabular}{c|cccccc}
\hline
ID & \begin{tabular}[c]{@{}c@{}}Size\\ $[-5, 10]$\end{tabular} 
   & \begin{tabular}[c]{@{}c@{}}Height\\ $[-10, 4]$\end{tabular} 
   & \begin{tabular}[c]{@{}c@{}}Squareness\\ $[-4, 20]$\end{tabular} 
   & \begin{tabular}[c]{@{}c@{}}Round-Shrink\\ $[-20, 20]$\end{tabular} 
   & \begin{tabular}[c]{@{}c@{}}Cat-Eye\\ $[-15, 20]$\end{tabular} 
   & \begin{tabular}[c]{@{}c@{}}Thickness\\ $[-10, 10]$\end{tabular} \\ \hline
1 & 3 & -3 & 7 & -2.5 & 3.5 & 6.5 \\
2 & -4 & -3 & 10 & -12 & -4.5 & 2.5 \\
3 & 6 & -3 & -1.5 & 2.5 & 1 & 5.5 \\
4 & -4 & 2 & -1 & 4 & 0 & -5 \\ \hline
\end{tabular}}
\caption{\textbf{Examples of Customized Eyewear:}
The top row shows four synthesized eyewear styles corresponding to IDs 1–4.
The table below lists the adjustable parameters used to generate each style.}
\label{fig:glasses}
\vspace{-2pt}
\end{figure}

\begin{table}[t]
\centering
\small
\caption{\textbf{Ablation Study on Temporal Smoothing:} 
We evaluate the stabilization effect by comparing the raw synthesized video (Before) with the result after applying temporal smoothing (After).}
\setlength{\tabcolsep}{18pt}{
\begin{tabular}{l|ccc}
\hline
Method & ITF$\uparrow$ & ISI$\uparrow$ & MOFM$\downarrow$ \\ \hline
Before & 32.2929 & 0.9284 & 0.4701 \\
After & 36.6289 & 0.9639 & 0.2101 \\ \hline
\end{tabular}}
\label{tab:ablation}
\vspace{-12pt}
\end{table}

\section{Method}
The overall framework of the proposed GlassesGB is shown in~\cref{fig:teaser}. 
Given a monocular video, we first perform facial alignment on all frames and crop them to $512\times512$ resolution, and then select a representative frame. 
By adjusting a set of glasses-style parameters (e.g., size, height, squareness, round shrink, cat-eye, and thickness) according to user preferences and the initial templates provided by GlassesGAN~\cite{plesh2023glassesgan}, we generate personalized eyewear. 
Examples of the parameter-adjusted results are shown in~\cref{fig:glasses}.

After selecting the preferred glasses, we apply temporal smoothing to synthesize all frames into a stable video. 
Specifically, the optical flow field $f_t$ between frame $t-1$ and frame $t$ is estimated using an optical-flow operator $\mathrm{Flow}(\cdot)$ applied to the GAN-edited frame pair $\left(I^{\mathrm{GAN}}{t-1}, I^{\mathrm{GAN}}{t}\right)$:
\begin{equation}
f_t = \mathrm{Flow}(I^{\mathrm{GAN}}_{t-1},\, I^{\mathrm{GAN}}_{t}).
\end{equation}
The output from the previous iteration, $I^{\mathrm{Final}}{t-1}$, is then warped toward frame $t$ according to $f_t$, producing:
\begin{equation}
\widetilde{I}_{t-1} = \mathrm{Warp}(I^{\mathrm{Final}}_{t-1},\, f_t),
\end{equation}
where $\mathrm{Warp}(I,f)$ performs pixel-wise remapping of $I$ using the flow field $f$. 
Finally, the temporally smoothed output frame is obtained with coefficient $\alpha$ as:
\begin{equation}
I^{\mathrm{Final}}_{t} = \alpha\, \widetilde{I}_{t-1} + (1 - \alpha)\, I^{\mathrm{GAN}}_{t}.
\end{equation}

Since glasses are not part of the FLAME~\cite{li2017learning} topology, we employ a facial tracker~\cite{zielonka2022towards} to estimate the neutral-expression FLAME mesh, the corresponding set of expression blendshapes, camera parameters $C$, joint and pose parameters $\Theta$, and expression coefficients $\psi_k$ from the input video. 
Linear Blend Skinning (LBS) is then applied to deform the Gaussian head model, which is subsequently rendered via Gaussian Splatting for real-time animation.

\section{Experiments}
To evaluate the effectiveness of the proposed GlassesGB framework, we conduct experiments on a video sourced from \href{https://www.youtube.com/watch?v=mKHgXHKbJUE}{YouTube}.

\textbf{Ablation Study on Temporal Smoothing:}
We employ three no-reference (NR) metrics to quantitatively assess the temporal smoothness of video sequences. 
Inter-frame Transformation Fidelity (ITF) computes the peak signal-to-noise ratio (PSNR, in~dB) between consecutive frames, reflecting the level of temporal consistency. 
Inter-Frame Similarity Index (ISI) measures the structural similarity (SSIM) between adjacent frames to evaluate perceptual coherence over time. 
Mean Optical Flow Magnitude (MOFM) captures the average motion intensity between successive frames, providing an estimate of overall temporal stability. 
From the results reported in~\cref{tab:ablation}, we can see that after applying temporal smoothing, all three metrics show significant improvement.

\begin{figure}[t]
\centering
\includegraphics[width=\linewidth]{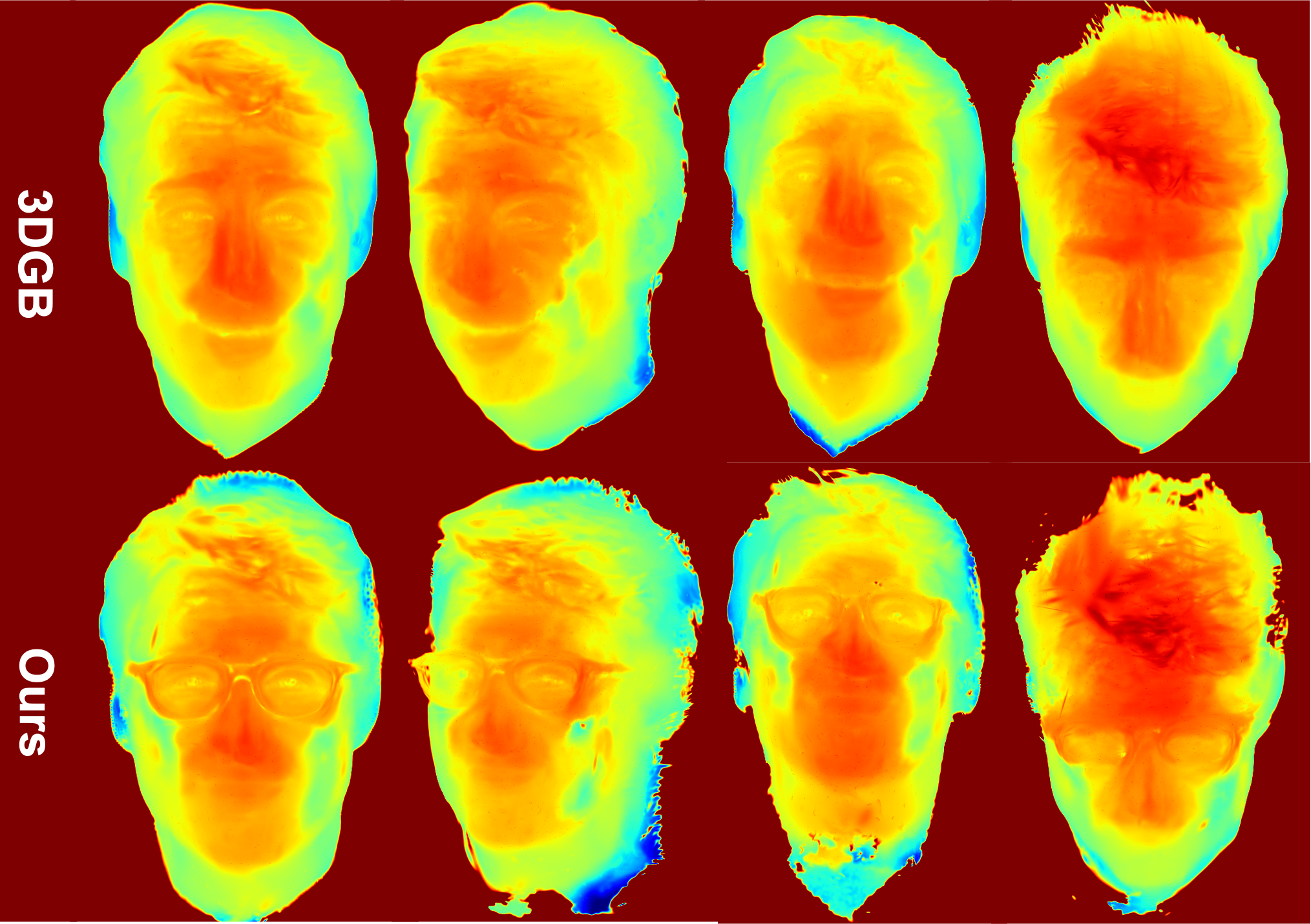}
\caption{\textbf{Depth Map Comparison:} 
We present pseudo-colored depth maps to show geometric details in the eyewear region.}
\label{fig:depth}
\vspace{-12pt}
\end{figure}

\textbf{Depth Map Comparison:}
We compare the depth maps of 3D head avatars synthesized by 3DGB~\cite{ma20243d} and our GlassesGB, as shown in~\cref{fig:depth}, where warmer colors denote regions closer to the camera and cooler colors correspond to greater depth.
The results show that the eyewear generated by GlassesGB exhibits real 3D geometry rather than appearing as texture overlays. 
Depth maps from novel views further demonstrate that GlassesGB preserves the original geometry of 3DGB while introducing additional structural details for the eyewear, without causing geometric degradation.

\section{Conclusion}
To enable VTON in VR scenarios, we propose GlassesGB, a novel framework that integrates GlassesGAN with 3DGB to render a photorealistic 3D head avatar with customized eyewear from a monocular input video.
In future work, we plan to further investigate VR deployment constraints such as stereo consistency, latency budgets, and headset-specific rendering artifacts.

\acknowledgments{This work was supported by the National Science and Technology Council of Taiwan, Grant Number: NSTC 114-2221-E-032-011-.}

\bibliographystyle{abbrv-doi}
\bibliography{main}
\end{document}